# Long-Range depth estimation using learning based Hybrid Distortion Model for CCTV cameras

Ami Pandat [1,*] 0000-0002-6882-9881, Punna Rajasekhar [2] 0000-0001-7132-6997 , G. Aravamuthan [2], Gopika Vinod [1,2] and Rohit Shukla [1,2]

[1] Homi Bhabha National Institute, Mumbai, Maharashtra-India, [2] Bhabha Atomic Research Center, Mumbai, Maharashtra-India
*Correspondence: a.pandat05@gmail.com

**Abstract:** Accurate camera models are essential for photogrammetry applications such as 3D mapping and object localization, particularly for long distances. Various stereo-camera based 3D localization methods are available but are limited to few hundreds of meters' range. This is majorly due to the limitation of the distortion models assumed for the non-linearities present in the camera lens. This paper presents a framework for modeling a suitable distortion model that can be used for localizing the objects at longer distances. It is well known that neural networks can be a better alternative to model a highly complex non-linear lens distortion function; on contrary, it is observed that a direct application of neural networks to distortion models fails to converge to estimate the camera parameters. To resolve this, a hybrid approach is presented in this paper where the conventional distortion models are initially extended to incorporate higher-order terms and then enhanced using neural network based residual correction model. This hybrid approach has substantially improved long-range localization performance and is capable of estimating the 3D position of objects at distances up to 5 kilometres. The estimated 3D coordinates are transformed to GIS coordinates and are plotted on a GIS map for visualization. Experimental validation demonstrates the robustness and effectiveness of proposed framework, offering a practical solution to calibrate CCTV cameras for long-range photogrammetry applications.

# 1. Introduction

Modern surveillance and automation systems increasingly depend on advanced sensing technologies for the detection and localization of distant objects in a wide range of applications, viz. border, maritime, aerial surveillance, and autonomous driving [1]. For example, localization of obstacles up to 300m is necessary in autonomous vehicles to ensure timely application of brakes and effective manoeuvring to avoid collisions [2]. In surveillance applications, it is required to detect fast moving objects such as drones at much longer distance (2-3kms) and often rely on active-sensing technologies such as LiDAR and RADAR. LiDAR is capable of producing highly detailed depth measurements in close to moderate ranges. However, its effectiveness declines rapidly with increasing distances [3]. RADAR-based systems have the ability to map terrain under various environmental conditions to generate 3D surface models and locate aerial objects. However, the spatial resolution of RADAR is lower compared to that of optical sensors. It faces significant challenges in detecting low-flying objects due to ground reflections and often fails to detect small objects such as drones due to their smaller radar cross sections [4]. Both these systems are expensive, highly power consuming, and require complex setups, making them less suitable for general-purpose or non-military deployments.

A ubiquitous and passive sensing technology that can be used for detection and 3D localization is an optical-based system [5], in which highly accurate state-of-the-art long-range object detection methods [6, 7] are already available. However, 3D reconstruction and 3D localization of distant targets still remain a challenge in this field. For example, an efficient real-time detector such as YOLO [ 8] reliably detects drones with sub-pixel accuracy; but accurate localization is limited to short ranges (8–10m) [9]. Various optical camera-based approaches such as Structure from Motion (SfM) [10 ,11] and stereo-vision have been explored for the reconstruction of 3D spatial information from 2D images. SfM reconstructs 3D structure and camera poses from uncalibrated image sequences, offering robustness to viewpoint changes, but at high computational cost. Stereo vision mimics human binocular perception and uses two spatially separated and calibrated cameras. These cameras are typically calibrated using Zhang's method of camera calibration [12]. The 3D coordinates are estimated from the corresponding/matching 2D image points in stereo images by triangulation [13] algorithm. This paper focuses on stereo vision due to its applicability in long-range photogrammetry and relevance to surveillance scenarios [14].

A pinhole camera model is an ideal and linear model that transforms the 3D coordinates of the world to 2D image points. However, this model does not represent the distortions in the image that arise due to manufacturing imperfections or asymmetric lens designs, which significantly degrades the photogrammetry estimates. In order to incorporate these non-linear characteristics of lens, non-linear distortion functions are introduced to model a realistic transformation. Expensive lenses with minimal distortion are recommended for long-range photogrammetry, while typical surveillance cameras often introduce radial and tangential distortions, especially near the edges of the images. Classical models such as Brown–Conrady [15], Kannala-Brandt [16], and rational models [17] are commonly used to model these effects. However, their ability to represent higher-order distortions and spatially varying effects is limited. Recent methods use end-to-end neural networks for distortion modeling [18] and intrinsic parameter estimation [19–21]. However, our study shows that neural network models alone do not converge for long-range photogrammetry tasks.

Incorrect lens models manifest large errors for long-range photogrammetry, and to address these issues, a novel hybrid distortion model is proposed in this paper. The hybrid model augments the classical distortion models with a learning-based residual correction. Integration of a learning-based module allows efficient correction as it incorporates the nonlinearities missed by the conventional models. The novel proposed method is capable of estimating 3D points at a distance of up to 5km. Experimental results demonstrate that this hybrid approach significantly improves localization accuracy even with CCTV surveillance cameras.

The remainder of this paper is organized as follows. Section 2 reviews related work on long-range photogrammetry and camera calibration. Section 3 discusses Zhang's method of camera calibration and triangulation algorithm. Section 4 describes the proposed framework in detail, including the stereo setup and the hybrid distortion model. Finally, Section 5 summarizes this paper.

## 2. Literature Survey

Over the years, numerous camera calibration techniques have been proposed to estimate the intrinsic and extrinsic parameters of the camera. An early attempt involved the design of an imaging model based on optimization algorithms [22], but required a complex computation and highly accurate initial values. The direct linear transformation (DLT) method [23] is another model that solves linear equations to determine camera parameters, although it does not account for lens distortion. A two-step method [24] that combines DLT with optimization is presented that addresses radial distortion only. A Zhang's planar calibration technique [12] estimates camera parameters using images of a planar pattern, i.e., checkerboard, captured from different views. By analysing the homography between the checkerboard and image planes, it accurately estimates camera parameters. However, the deployment of these physical targets becomes impractical for long-range or wide-field-of-view systems in outdoor setups. Self-calibration methods overcome this limitation by exploiting natural scene features and camera motion to infer calibration parameters without predefined targets [25–27]. However, this method does not work well in environments such as the sky, desert, or sea, where there are few distinctive features. A method based on active vision was suggested in [28,29], which requires the camera to move in a specific way, which may not be suitable for cases where the cameras are fixed and the field-of-view is large. Several other studies have also addressed calibration methods for wide field-of-view scenarios. For example, GPS devices are used as feature points [30], but this limits the flexibility of the approach. A stereo-based 3D measurement system using a cross-shaped target with ring-coded markers was also used [31]. A method in which the optical center of the camera and the target control points are nearly in the same plane [32] is proposed, but it neglects longitudinal tangential distortion, which can become significant over long distances. Another calibration approach for large field-of-view was introduced [33], where the target is positioned randomly within the field-of-view of the camera, several times. However, this method requires at least five 3D calibration points at long distances to achieve accurate calibration, which is impractical in real-world deployments. Another stereo calibration method tailored for out-of-focus cameras extracts feature points through ellipse fitting using active phase-shift circular grating (PCG) arrays [34,35]. However, it is less suitable for long-range photogrammetry because of challenges in detecting patterns at long distances.

Among these, Zhang's method is widely adopted and has been a standard approach in stereobased depth estimation. It is widely applied in various domains such as autonomous vehicles [36], robotics, forest environment mapping [37], underwater object localization [38] etc. Recent studies show that with appropriate hardware and using Zhang's calibration techniques, stereo vision systems typically achieve effective depth estimation between 1.5m and 300m. Zhang's method was used for depth estimation up to 200m [36] in autonomous driving. In agricultural applications, a 20cm baseline stereo with Zhang's calibration is used to localize crops such as tomatoes and potatoes in shorter ranges [39,40]. Several studies have aimed to improve Zhang's calibration technique for long-range depth estimation. For example, super-resolution reconstruction of the original image is used to localize ships up to 300m [41] and UAV-based localization in a low-altitude environment achieves up to 100m [42]. A recent advanced photogrammetry method [43] that uses Zhang's method and high-quality lenses assuming zero distortion has reported depth estimation capabilities at 1km, but limited to a radius of 100m. It is clear from this existing literature that stereo-based depth estimation is limited to the few hundreds of meters.

One of the factors affecting the long-range estimation in Zhang's method is lens distortion models. Most stereo depth estimation systems use distortion models with only 3 to 5 parameters, inherently limiting their effective range to approximately 100–300 meters [44–46]. Our previous work [47] also adopts a five-parameter distortion model in a monocular setup, which assumes that the targets are located on a known ground plane and are at a distance of 100m. A reliable localization beyond 100m requires high-resolution imagery, meticulous calibration, and often large baselines. Another factor affecting estimation is the disparity error, and the study [48] shows that even a 0.5-pixel disparity error manifest as large errors at long distances.

This paper emphasizes the need for more accurate and detailed distortion modeling, along with sub-pixel level detection for long-range photogrammetry. The novel hybrid distortion model proposed in this study is designed for medium-quality surveillance cameras and aims to deliver accurate depth estimation up to 5km, a significant improvement over traditional models using similar hardware.

## 3. Camera Calibration and Stereo-Localization

Camera calibration is the process of estimating camera parameters and, as discussed in section 2, Zhang's method [12] is widely used for calibration. It uses a planar target (for example, a checkerboard) captured from multiple views and estimates the intrinsic and lens distortion parameters, along with the extrinsic parameters for each view by minimizing the reprojection error. These parameters are used to localize objects in the 3D space using triangulation methods.

### 3.1 Camera Model and Parameters

Assuming the pinhole model for the camera, the linear projection of a Euclidean point $\mathbf{X_e} = (X, Y, Z)^T$ in 3D world coordinates to a 2D point $\mathbf{U_e} = (u, v)^T$ in the Euclidean image plane is given by the direct linear transform.

Let $\mathbf{X} = (X, Y, Z, 1)^T$ be the representation of $\mathbf{X_e}$ in homogeneous coordinate system; similarly, $\mathbf{U} = (u, v, 1)^T$ for $\mathbf{U_e}$; $\mathbf{P} = (x, y, 1)^T$ be the projection of a 3D point $\mathbf{X_e}$ in camera coordinate system. Assuming a pinhole model, these are linearly related as follows:

$$\mathbf{P} = \lambda [\mathbf{R} \quad \mathbf{t}] \mathbf{X} \quad (1)$$

$$\mathbf{U} = \mathbf{A}\mathbf{P} \quad (2)$$

where:
- $\lambda$ is a scale factor.
- **R** is the **3x3** rotation matrix and **t** is the **3x1** translation vector defining the camera pose w.r.t the 3D world coordinate system, together represented as *W*.
- **A** is **3x3** intrinsic matrix given by,

$$A = \begin{bmatrix} f_x & \gamma & c_x \\ 0 & f_y & c_y \\ 0 & 0 & 1 \end{bmatrix} \quad (3)$$

where $f_x, f_y$ represent focal lengths in $x$ and $y$ directions, respectively, $c_x, c_y$ as principal point coordinates and $\gamma$ is the skew.

## 3.2 Distortion Model

The transformation given in Eq. 2 describes a linear relation between the 3D and 2D planes. In real cameras, distortions are introduced by the camera lens and these distortions are modelled as the sum of multiple non-linear distortion functions, with following parameters:

- $k_1, k_2, k_3$ are **Radial distortion parameters**
- $p_1, p_2$ are **Tangential distortion parameters**
- $k_4, k_5, k_6$ are **Rational model distortion parameters**
- $s_1, s_2, s_3, s_4$ are **Thin prism distortion parameters**
- $\tau_x, \tau_y$ are **Tilted sensor distortion parameters** which used to form a 3x3 transformation matrix, $M_\varnothing$

All the parameters in the distortion models are represented as a vector **K**.

The linear camera model in Eq.1 is modified in most implementations, as shown in Eq.4.

The projected point $\mathbf{P} = (x, y, 1)^T$ is distorted to $\mathbf{P_D} = (x_D, y_D, 1)^T$ through an intermediary step $\mathbf{P_I} = (x_I, y_I, 1)^T$

$$\begin{bmatrix} x_I \\ y_I \end{bmatrix} = \frac{(1 + k_1 r^2 + k_2 r^4 + k_3 r^6)}{1 + k_4 r^2 + k_5 r^4 + k_6 r^6} \begin{bmatrix} x \\ y \end{bmatrix} + \begin{bmatrix} 2p_1 xy + p_2(r^2 + 2x^2) \\ p_1(r^2 + 2y^2) + 2p_2 xy \end{bmatrix} + \begin{bmatrix} s_1 r^2 + s_2 r^4 \\ s_3 r^2 + s_4 r^4 \end{bmatrix} \quad (4)$$

$$\mathbf{P_D} = \mathbf{M_\varnothing P_I}$$

$$\mathbf{U} = \mathbf{A P_D}$$

where $r = \sqrt{x^2 + y^2}$ is the radial distance from the optical center.

## 3.3 Calibration Procedure

An outline of the Zhang's camera calibration algorithm [12] is summarized in Algorithm 1.

### 3.3.1. Reprojection Error

Reprojection error represents the discrepancy between the observed image points (i.e., the points detected in the image of the calibration object) and the projected points (i.e., the points calculated using the estimated camera parameters). The total reprojection error for a calibration procedure involving $M$ images and $N$ points per image is defined as the sum of the reprojection errors for all observed points.

$$E_{RMS} = \sqrt{\frac{1}{MN} \sum_{i=1}^{M} \sum_{j=1}^{N} \|x_{ij} - \hat{x}_{ij}\|^2} \quad (5)$$

Where:

- $x_{ij}$ is the observed 2D image point of the $j^{th}$ 3D world point in the $i^{th}$ image.
- $\hat{x}_{ij}$ is the corresponding projected 2D point using the estimated camera intrinsics and extrinsics.
- $\|\cdot\|$ denotes the Euclidean norm (typically L2).

A good calibration generally yields a reprojection error of less than 1 pixel. So, the camera calibration process aims to find the set of intrinsic parameters **A**, extrinsic parameters **R**, **t**, and distortion coefficients **K** that minimizes the total reprojection error to 0 [49].

---

**Algorithm 1: Zhang's Camera calibration algorithm (overview)**

It is assumed that each observed image point $\mathbf{U}_{ij}$ corresponds to the associated model point $\mathbf{X}_j$, and all views are from the same camera.

**Input** : $\mathcal{X} = (\mathbf{X}_0, \ldots, \mathbf{X}_{N-1})$, an ordered sequence of 3D points on the planar target, with $\mathbf{X}_j = (x_j, y_j, 0)^\top$;
$\mathcal{U} = (\mathbf{U}_0, \ldots, \mathbf{U}_{M-1})$, a sequence of views, each view $\mathbf{U}_i = (u_{i0}, \ldots, u_{i,N-1})$, an ordered sequence of image points $u_{ij} = (u_{ij}, v_{ij})^\top$

**Output**: Intrinsic parameters **A**, **K**; extrinsic parameters $\mathbf{W} = (\mathbf{W}_0, \ldots, \mathbf{W}_{M-1})$, where $\mathbf{W}_i = (\mathbf{R}_i, \mathbf{t}_i)$

1: $\mathbf{H}_{init} \leftarrow \text{GETHOMOGRAPHIES}(\mathcal{X}, \mathcal{U})$; where Homography is a 3x3 matrix which transforms 2D points from one plane to another.
2: $\mathbf{A}_{init} \leftarrow \text{GETCAMERAINTRINSICS}(\mathbf{H}_{init})$
3: $\mathbf{W}_{init} \leftarrow \text{GETEXTRINSICS}(\mathbf{A}_{init}, \mathbf{H}_{init})$
4: $\mathbf{K}_{init} \leftarrow \text{ESTLENSDISTORTION}(\mathbf{A}_{init}, \mathbf{W}_{init}, \mathcal{X}, \mathcal{U})$
5: $(\mathbf{A}, \mathbf{K}, \mathbf{W}) \leftarrow \text{REFINEALL}(\mathbf{A}_{init}, \mathbf{K}_{init}, \mathbf{W}_{init}, \mathcal{X}, \mathcal{U})$; to optimize camera parameters using Levenberg Marquardt optimization.
6: **return** $(\mathbf{A}, \mathbf{K}, \mathbf{W})$

---

### 3.3.2. Optimization

The problem of minimizing the reprojection error for the above system of non-linear functions defined in Eq. 4 can be solved using optimization algorithms such as Levenberg–Marquardt (LM) [50], Gradient Descent (GD) and its variants such as ADAM.

The Levenberg–Marquardt update equation used in camera calibration is given by:

$$\theta_{new} = \theta + (\mathbf{J}^\top \mathbf{J} + \lambda \mathbf{I})^{-1} \mathbf{J}^\top \mathbf{r} \tag{6}$$

where:

- $\theta$ is the current estimate of the camera calibration parameters
- **J** is the Jacobian matrix of the residuals with respect to $\theta$,
- $\lambda$ is the damping term,
- **I** is the identity matrix,
- **r** is the residual vector, defined as the difference between the observed image points and the projected points computed using the current parameters.

$$\mathbf{r} = \mathbf{y} - f(\theta) \tag{7}$$

where **y** are observed 2D image points and $f(\theta)$ denotes the projection of 3D world points onto the image plane using the current estimate of the calibration parameters (**A**, **K**, **W**). GD and its variants are first-order methods that are not commonly employed in any major implementation

## 3.4 3D Localization Method: Triangulation

Triangulation is the process of determining the 3D location of a point by observing it from multiple views. In stereo-vision, this involves computing the 3D coordinates of a point seen in both the left and right camera images [13]. The core principle relies on projective geometry, where a 3D point in the world projects to 2D image coordinates in each camera view. Given the projection matrices of the left and right cameras, $P_L$ and $P_R$, and the corresponding image points of each view, the objective is to find the most likely 3D point **X** that explains these observations.

The projection matrices of the left and right cameras $P_L$ and $P_R$, respectively, are obtained from the calibration of each camera after correction for distortion.

$$P_L = A_L W_L \\ P_R = A_R W_R \qquad (8)$$

**Where:**

- $A_L$ and $A_R$ are the intrinsic matrices for left and right cameras
- $W_L$ and $W_R$ are the extrinsic matrices of size $3 \times 4$ for left and right cameras, evaluated based on a local reference system, called world coordinate system.

Let $X$ be the 3D point in homogeneous world coordinates, and the image coordinates represented in homogeneous coordinates be $U_L = [u_L, v_L, 1]^T$, and $U_R = [u_R, v_R, 1]^T$. To determine the 3D point $X = [X, Y, Z, 1]^T$, the system of projection equations $U_L = P_L X$ and $U_R = P_R X$ needs to be solved.

These can be rewritten as a homogeneous linear system $HX = 0$. Here, $H$ is

$$H = [u_L P_L^3 - P_L^1 \quad v_L P_L^3 - P_L^2 \quad u_R P_R^3 - P_R^1 \quad v_R P_R^3 - P_R^2]^T \qquad (9)$$

where $P_k^i$ denotes the $i^{th}$ row of the matrix $P_k$.

This system is solved as a constrained optimization problem, assuming $X^T X = 1$, whose solution is derived from Singular Value Decomposition (SVD) of $H$. The singular vector corresponding to the minimum singular value is the solution for $X$. This yields the 3D location of the object relative to the chosen world coordinate system. If the stereo rig's GPS position is known, this local coordinate can be further transformed into GIS coordinates (latitude and longitude) for mapping.

## 4. Experiments and Analysis

This section describes the implementation and results of 3D object localization at long-range using a stereo camera setup. The setup consists of two surveillance cameras with a baseline distance of 10 meters. The concept of stereo triangulation is illustrated in Fig.4. The checkerboard indicates the common scene points used to find the extrinsic parameters of the cameras.

### 4.1 Camera calibration

Estimation of camera parameters through camera calibration is the first step in 3D localization which are further used by triangulation method discussed in Section 3.4. The Zhang's method (as described in Section 3) was used to calibrate both cameras independently. The data required for calibration are obtained by capturing multiple images of a checkerboard with dimensions 6×7 and a square size of 5 cm. As illustrated in Fig.1, these images were captured to cover most of the area within the field-of-view of the camera.

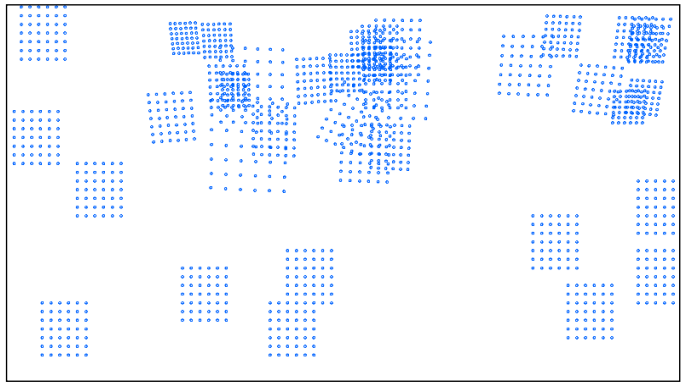

**Figure 1.** Representative image showing the area covered by checkerboard in multiple views during Camera Calibration. Each set of dots indicates the position of the checkerboard in a view.

The calibration process estimates the intrinsic matrix **A**, and distortion parameters **K** for each camera. Usually, the computer vision research community uses OpenCV [51] for camera calibration due to their extensive support for vision algorithms. OpenCV estimates up to 14 distortion parameters of various distortion models, including radial, rational, tangential, thin-prism, and tilted sensor models. The combination of all these models forms a distortion vector **K**. Fig.2 shows the standard pipeline to project 3D world points onto a 2D image plane. A calibration framework consisting of the distortion model described by Eq. 4 is implemented in PyTorch, to study distortion models. The correctness of the framework is verified by comparing the calibration outputs of various cameras with OpenCV. The comparison of the reprojection errors is shown in Table 1 and it can be seen that they closely match the OpenCV implementation.

## 4.2 Investigation of Lens Quality

The influence of lens quality on reprojection error can be seen from the Table 1. Specifically, surveillance camera lenses exhibited substantially higher reprojection errors (approximately 0.2 pixels) compared to machine vision lenses, whose reprojection error is around 0.07 pixels, three times lower.

**Table 1.** Comparison of reprojection Errors obtained in our implementation and OpenCV for multiple cameras using 14-parameter distortion model

| Camera | OpenCV | Our implementation |
| --- | --- | --- |
| Surveillance Camera | 0.22 | 0.20 |
| Surveillance Camera | 0.24 | 0.21 |
| Machine Vision Lens | 0.07 | 0.07 |
| Webcam | 0.10 | 0.08 |

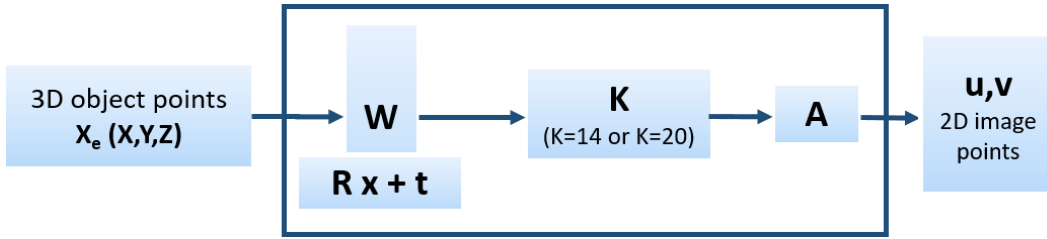

**Figure 2.** Camera Calibration pipeline

A similar study [43] reported minimal reprojection error, as low as 0.08 pixels, using machine vision telephoto lenses for long-range photogrammetry; however, their approach did not explicitly account for lens distortion, a factor that can significantly affect localization accuracy. Although high quality lenses allowed for accurate localization of objects at distances of up to 1 km, it was limited to a radius of 100m. Their approach did not include comprehensive distortion corrections [43], limiting practical usage to long range surveillance scenarios. These limitations motivated this study to improve the lens distortion models in the surveillance CCTV camera.

**Table 2.** Depth Error ($\Delta Z$) at Varying Disparities and Distances

| $\Delta d$ (pixel) | $\Delta Z$ (z = 1 km) | $\Delta Z$ (z = 5 km) |
| --- | --- | --- |
| 2 | 125 m | 3.125 km |
| 1 | 62.5 m | 1.5625 km |
| 0.5 | 31.25 m | 781.25 m |
| 0.1 | 6.25 m | 156.25 m |

## 4.3 Lens modeling for long-range photogrammetry

In the stereo setup described in Section 4.1, the camera parameters of both cameras are estimated through calibration on a world-coordinate system, which is common to both cameras. These parameters can now be used to estimate the 3D coordinates of the corresponding image points (test points) in both images. For this analysis, 7 test points are picked from the images, which correspond to world points at varying distances from 100m up to 5km from the origin (where the cameras are placed). These test points are input to the triangulation algorithm and needs to be sub-pixel accurate. As shown in Table 2, for a pinhole camera model with a simplistic parallel-stereo setup, even a small error of 0.1 pixels can result in a depth estimation discrepancy of up to 156.25 meters at a distance of 5km. So, it is required to input sub-pixel accurate corresponding points for localization. To obtain sub-pixel accurate point correspondences, a simple up-sampling based heuristic approach has been used. First, small patches containing the points-of-interest are cropped from the original high-resolution image.

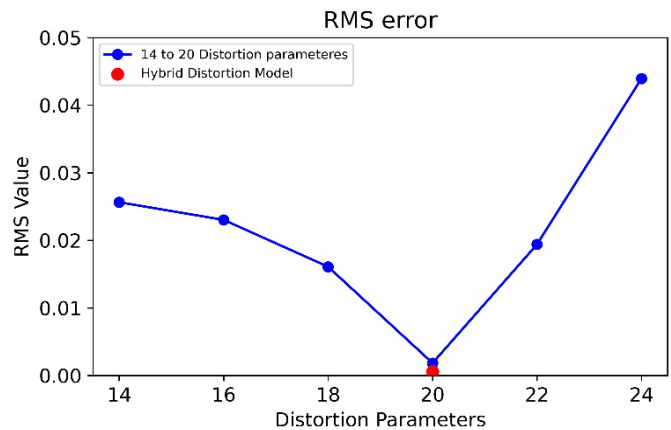

**Figure 3.** RMS Error between evaluated and original GIS Coordinates of Test Points

These patches are then up-scaled by a factor of 10 using bilinear interpolation, allowing us to manually mark the corresponding points with higher precision in the enlarged images. After marking, the obtained corresponding points were scaled down by 10 to recover sub-pixel values relative to the original cropped patch. Finally, the crop offset is added to map these sub-pixel coordinates to the full-resolution image. As shown in Fig.6(**a**) and Fig.6(**b**), integer pixels have higher depth error than sub-pixel points. In Fig.6(**a**), the projected points belonging to 5 and 6 are out of the scene, but in Fig.6(**b**) they are within in the scene. Subsequently, the 3D coordinates for each of these 7 test points will be estimated using the triangulation algorithm. To verify the accuracy of the 3D location estimates for these 7 test points, the coordinates will be transformed into latitude and longitude using the Geopy Python library, which utilizes the WGS84 [52] coordinate system. Various lens distortion models studied are presented in this section.

### 4.3.1. Neural Network based distortion model
The first attempt to modify the lens distortion model of a medium-quality surveillance camera involved reformulating the distortion vector **K** as a single neural network. This is motivated by the established use of neural networks as flexible approximators in non-linear function modeling.

As illustrated in Fig.5, the distortion vector **K** was replaced with a 4-layered neural network block. The network was implemented in PyTorch [53]and the parameters are optimized using the ADAM optimizer [54]. The parameters were observed to diverge, and this can be attributed to the optimizer. The same setup tested with other gradient descent based optimizers too. It has been observed that any first-order methods such as Gradient Descent and its variants are not suitable for optimizing models with fewer parameters and work best with complex models with a large number of parameters. On the other hand, the LM optimizer works well for a simpler model with a small set of parameters such as the camera calibration model [55]. This experiment suggests that a standalone neural network is not suitable as a non-linear lens distortion model.

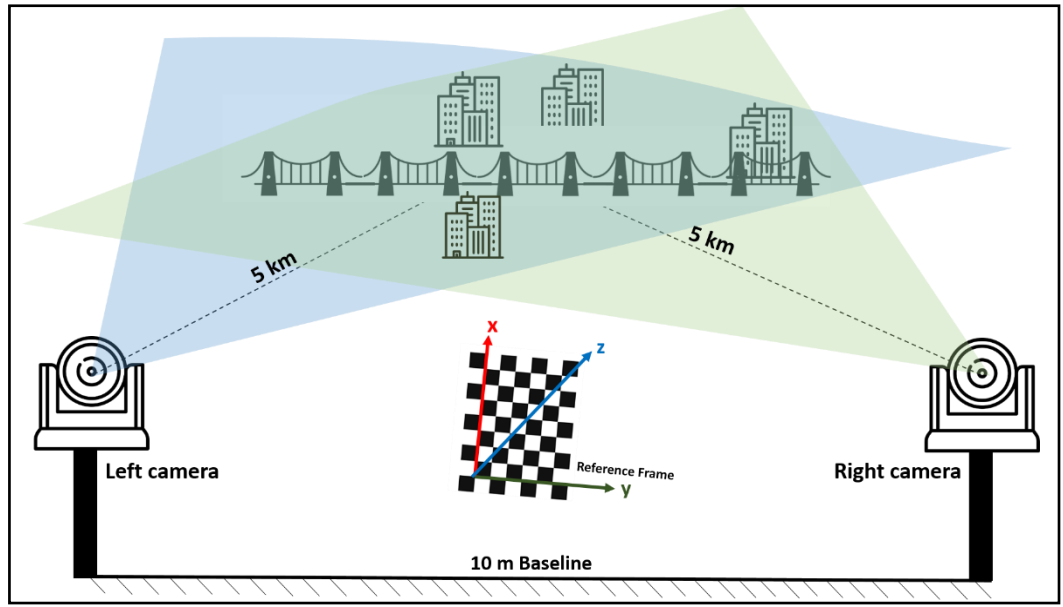

**Figure 4.** Stereo Camera setup for 3D Triangulation uses two surveillance camera named as left camera and right camera. Shaded region shows common field-of-view between two cameras

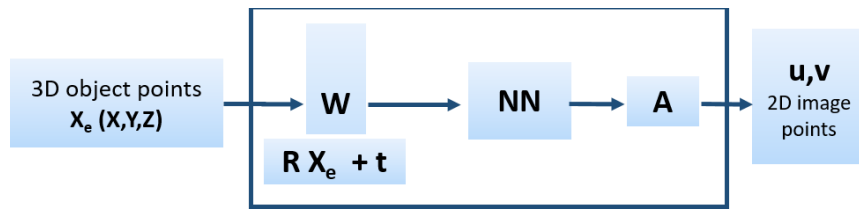

**Figure 5.** Camera Calibration pipeline with Neural Network based lens model

### 4.3.2. Extended distortion model

As standalone neural network based lens distortion model failed to converge, the classical 14 parameter distortion model as shown in Eq.4 is used for calibration. The parameters are estimated by camera calibration and 3D coordinates of all 7 test points are estimated.

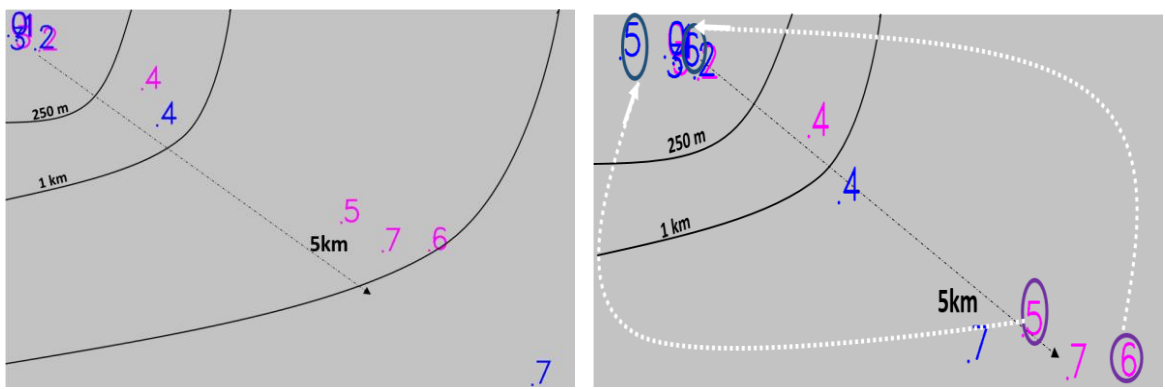

(**a**) Test points marked manually (Integer pixels)   (**b**) Sub-pixel accurate input to Triangulation

**Figure 6.** Illustration of Estimated Latitude Longitude with original values. The points marked in magenta colour are original true values and points marked in blue color are estimated values. The left image is test points marked manually (in integer pixels) and the right image is with test points marked in sub-pixel using heuristic-based approach. In right image, points are more accurate than left image i.e, when marked manually. It can be seen that beyond 250m arc, the estimates of the classical distortion model are inaccurate. The estimates of points 5 and 6, which are near 5km arc, are diverged from north-east to south-west direction (white arrow). This shows the limitation of classical distortion model with integer pixel corresponding points at long ranges.

These 3D coordinates are then converted to GIS coordinates (latitude and longitude) and mapped. It has been observed that GIS localization estimates are accurate up to 250m from origin and then diverge significantly beyond that. The test points with their original and estimated latitude and longitude values in Fig.6(**b**) have been illustrated. It shows the limitation of the classical 14-parameter distortion model. The maximum reprojection error evaluated for these 7 test points is 10.62 pixels as shown in Fig.9.

Hence, the 14parameter classical distortion model is found to be unsuitable for ranges beyond 250m. To overcome this, the 14-parameter distortion model is extended to include higher-order terms, and the number of distortion parameters is systematically increased from 14 to 24 as shown in Eq. 10 by incorporating higher-order terms.

$$\begin{bmatrix} x_p \\ y_p \end{bmatrix} = \frac{(1 + k_1 r^2 + k_2 r^4 + k_3 r^6 + k_7 r^8 + k_8 r^{10})}{1 + k_4 r^2 + k_5 r^4 + k_6 r^6 + k_9 r^8 + k_{10} r^{10}} \begin{bmatrix} x \\ y \end{bmatrix} + \begin{bmatrix} 2p_1 xy + p_2(r^2 + 2x^2) \\ p_1(r^2 + 2y^2) + 2p_2 xy \end{bmatrix} + \begin{bmatrix} s_1 r^2 + s_2 r^4 + s_5 r^6 \\ s_3 r^2 + s_4 r^4 + s_6 r^6 \end{bmatrix}$$
(10)

**Where:**

- $k_1, k_2, k_3, k_7, k_8, k_4, k_5, k_6, k_9, k_{10}$ are **Radial distortion coefficients**
- $p_1, p_2$ are **Tangential distortion coefficients**
- $s_1, s_2, s_3, s_4, s_5, s_6$ are **Thin prism distortion coefficients**
- $\tau_x, \tau_y$ are **Tilted sensor distortion parameters** which used to form a 3x3 transformation matrix

Table 3. Reprojection Errors for Various Distortion Parameters

| Sr. No | Parameters | Right Camera | Left Camera |
|---|---|---|---|
| 1 | 14 | 0.2021 | 0.2132 |
| 2 | 16 | 0.2020 | 0.2129 |
| 3 | 18 | 0.2021 | 0.2129 |
| 4 | 20 | 0.2019 | 0.2129 |
| 5 | 20 | 0.2018 | 0.2126 |
| 6 | 22 | 0.2017 | 0.2126 |
| 7 | 24 | 0.2017 | 0.2123 |

For each of these models, the optimized camera parameters are estimated and the 3D coordinates of the checkerboard image points used for calibration are obtained. The 3D coordinates obtained are reprojected on the image plane, and the reprojection errors are evaluated between the original and reprojected points for each of these models. The results are shown in Table 3.

The RMS error between the original and estimated GIS coordinates is calculated for 7 test points also for all models, and the results are plotted in Fig.3. For the current set of cameras, the RMS error reaches a minimum at 20 parameters model (Eq. 10) and increases beyond it, indicating that the model begins to overfit. Therefore, a distortion model consisting of 20 parameters as defined in Eq. 10, was selected for this stereo pair. Our framework can be used to automate the process of finding the right combination of parameters for a given camera pair from the minimal reprojection error on test points. With the extended distortion model of 20 parameters, the maximum reprojection error of the 7 test points evaluated was found to be 7.21 pixels as shown in Fig.9 which significantly lower than standard 14 parameter distortion model.

**4.3.3. Enhanced hybrid distortion model**

In order to further reduce the reprojection errors, the extended model is enhanced with a learning based residual correction for the intrinsic parameters. For this a neural network(NN) with three layers trained to estimate additive residual adjustments. These residual adjustments are added to the previously estimated intrinsic parameters, as shown in Eq. 11.

$$A = A^* + \delta_A$$
$$K = K^* + \delta_K \qquad (11)$$

Where,
- $A^*$ is the intrinsic matrix derived from 20 parameter model
- $K^*$ is the distortion parameter vector derived from 20 parameter model
- $\delta_A$, $\delta_K$ are learned residuals (output of neural network)

Fig.8 presents proposed hybrid model approach. Modules added to enhance the extended 20parameter model are highlighted with red lines. The neural network was trained with output of 20-parameter distortion model($K^*$) as input to correct residual nonlinearities. The output of the trained network is residuals that are added to the intrinsic parameters ($A^*$ and $K^*$). As shown in Fig.8, an inverse neural network is also trained along with it, which will be used to undistort the points, prior to triangulation. MSE loss (Eq. 12) is used to train the neural network as shown in Eq. 12.

$$L_{\text{mse}} = \text{MSE}(y_3, y_3^*) + \lambda\, \text{MSE}(y_1, y_4) \qquad (12)$$

**Where:**

- $y_3$ is the Original 2D image points
- $y_3^*$ is the estimated 2D image points
- $y_1$, is the input to Neural Network
- $y_4$ is the output of the Inverse Neural Network block

The maximum reprojection error across the 7 test points was found to be 6.93 pixels. The comparison of the standard-14-parameter, extended-20-parameter and hybrid learning-based distortion model is illustrated in Fig.9. As the distance of the object increases, the reprojection error gradually increases across all the three models described here, and the hybrid distortion model outperforms both the standard 14-parameter distortion model and the 20-parameter distortion model.

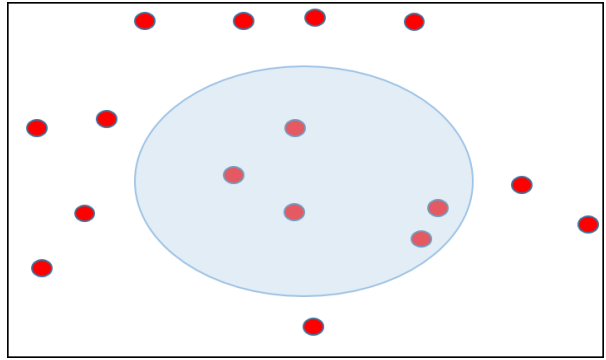

**Figure 7.** Illustration of Center and Edge pixels in an image. Shaded region represents center area of an image.

The distortion effect on test points can be further analysed based on their spatial placement within the image. As illustrated in Fig.7, the shaded region represents the center of the image, while the remaining points are located near the edges. Edge pixels are more susceptible to radial distortion than points that lie at the center of an image. Although increasing the number of polynomial terms in the distortion model helps mitigate these effects, the reprojection error remains higher at the edges than at the center as shown in Fig.10.

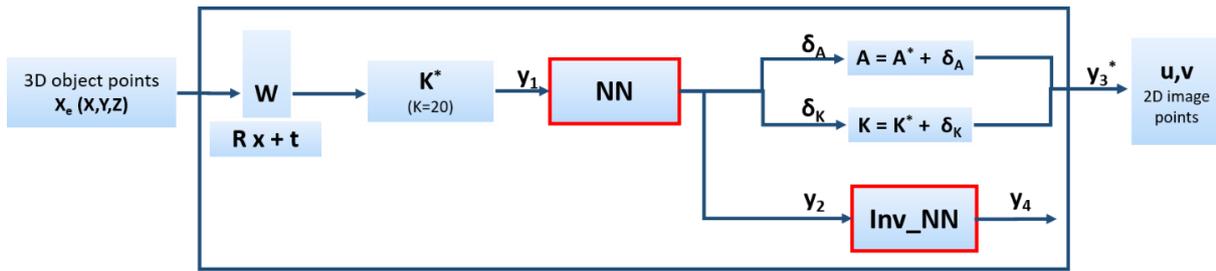

**Figure 8.** Hybrid distortion model. NN represents neural network block modeling the distortion and Inv_NN represents inverse neural network modeling the undistortion.

The estimated GIS coordinates using the 20-parameter model and the hybrid model are illustrated in Fig.11(a) and Fig.11(b) respectively. It can be seen that the hybrid model performs effectively beyond 1km and up to 5km, compared to the output of classical distortion model shown in Fig.6(b).

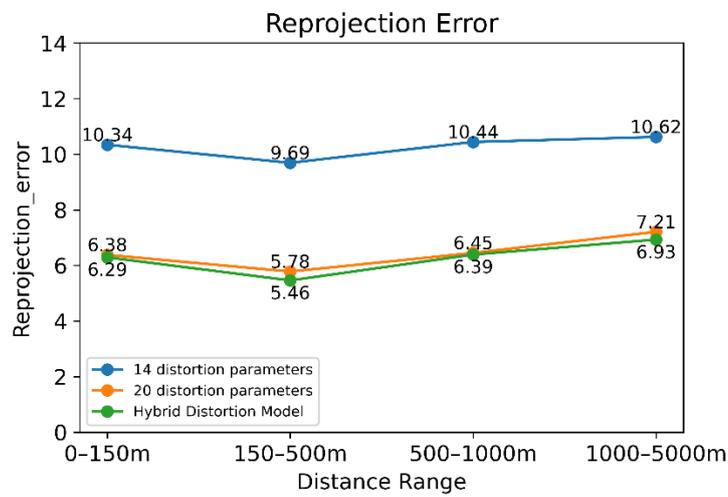

**Figure 9.** Comparison of Reprojection Error for various distortion models as the distance increases.

The lower reprojection error at 150-500m in Fig.9 is due to the test points being at center in that range, emphasizing the lower distortion effect than edge pixels. This trend underscores that, for long-range photogrammetry, the hybrid distortion model offers superior performance in minimizing reprojection error.

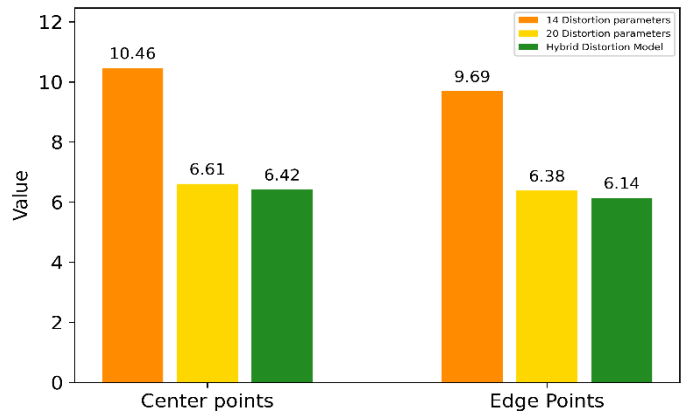

**Figure 10.** Reprojection Error of various models categorised based on their spatial location in image. Test points lie at the center of the image are center points and points near edges are edge points

The residual correction provided by the neural network further enhances the performance of the classical 20-parameter distortion model at the 5km range. In particular, Fig.11(**b**) shows that points 6 and 7 move closer to their original locations.

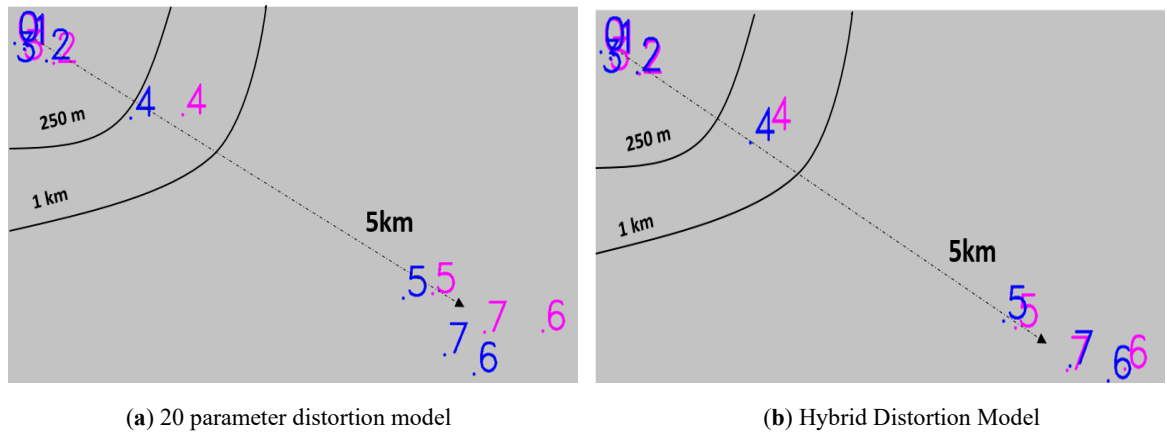

(**a**) 20 parameter distortion model  (**b**) Hybrid Distortion Model

**Figure 11.** Latitude and longitude Illustration of Test points. The points marked in magenta color are original true values and points marked in blue color are estimated values. It can be seen that the estimates of the hybrid distortion model outperform 20 parameter distortion model.

Both numerically (through reprojection error) and visually (via test point illustrations), it has been observed that this hybrid distortion model which combines a 20-parameter distortion model with neural network-based correction significantly improves ground-based stereo calibration for long-range photogrammetry.

## 5. Conclusion

Reliable 3D object localization is essential for applications in surveillance, autonomous navigation, and smart transportation systems. Among the various modalities explored in prior work such as LiDAR, RADAR, UAV-based sensing, and stereo-vision; ground-based stereo setups remain among the most practical and cost-effective solutions. However, existing stereo methods are generally limited to operating ranges of less than 250 meters. In this study, a stereo-vision based system designed for long range photogrammetry (up to 5 kilometres) is proposed. Using two identical surveillance cameras with a 10m baseline, Zhang's calibration technique employed to estimate intrinsic and extrinsic parameters, achieving reprojection errors around 0.2 pixels. The analysis revealed that lens quality significantly affects system range and performance. To address this, this paper introduced a novel hybrid distortion model, an extension of the classical distortion model enhanced with learning-based refinement of the intrinsic parameters. This learning-based residual correction led to more reliable localization at long distances. The results show that by incorporating the hybrid distortion model into the calibration pipeline, even low-cost CCTV surveillance cameras can be leveraged for long-range 3D localization, across a range of vision-based monitoring applications.

## Ethics Statement

This study did not involve human participants, animal subjects, or sensitive data requiring ethics approval.

## References

1. Faseeh, M.; Bibi, M.; Khan, M.A.; Kim, D.H. Deep learning-assisted real-time object recognition and depth estimation for enhancing emergency response in an adaptive environment. *Results in Engineering* **2024**, *24*, 103482. https://doi.org/https://doi.org/10.1016/j.rineng.2024.103482.

2. Khoche, A.; Sánchez, L.P.; Batool, N.; Mansouri, S.S.; Jensfelt, P. Towards Long-Range 3D Object Detection for Autonomous Vehicles. In Proceedings of the 2024 IEEE Intelligent Vehicles Symposium (IV), 2024, pp. 2206–2212. https://doi.org/10.1109/IV55156.2024.10588513.
3. Kim, J.; Park, B.j.; Kim, J. Empirical Analysis of Autonomous Vehicle's LiDAR Detection Performance Degradation for Actual Road Driving in Rain and Fog. *Sensors* **2023**, *23*. https://doi.org/10.3390/s23062972.
4. Khan, M.A.; Menouar, H.; Eldeeb, A.; Abu-Dayya, A.; Salim, F.D. On the Detection of Unauthorized Drones—Techniques and Future Perspectives: A Review. *IEEE Sensors Journal* **2022**, *22*, 11439–11455. https://doi.org/10.1109/JSEN.2022.3171293.
5. Schultheiss, P.M.; Wagner, K., Active and Passive Localization: Similarities and Differences. In *Underwater Acoustic Data Processing*; Springer Netherlands: Dordrecht, 1989; pp. 215–232. https://doi.org/10.1007/97894-009-2289-1_23.
6. Wang, H.; Mou, X.; Mou, W.; Yuan, S.; Ulun, S.; Yang, S.; Shin, B.S. Vision based long range object detection and tracking for unmanned surface vehicle. In Proceedings of the 2015 IEEE 7th International Conference on Cybernetics and Intelligent Systems (CIS) and IEEE Conference on Robotics, Automation and Mechatronics (RAM), 2015, pp. 101–105. https://doi.org/10.1109/ICCIS.2015.7274604.
7. Cheng, G.; Yuan, X.; Yao, X.; Yan, K.; Zeng, Q.; Xie, X.; Han, J. Towards Large-Scale Small Object Detection: Survey and Benchmarks. *IEEE Transactions on Pattern Analysis and Machine Intelligence* **2023**, *45*, 13467–13488. https://doi.org/10.1109/TPAMI.2023.3290594.
8. Redmon, J.; Divvala, S.; Girshick, R.; Farhadi, A. You Only Look Once: Unified, Real-Time Object Detection, 2016, [arXiv:cs.CV/1506.02640].
9. Sharma, A.; Jain, N.; Kothari, M. Lightweight Multi-Drone Detection and 3D-Localization via YOLO, 2022, [arXiv:cs.CV/2202.09097].
10. Nielsen, M.S.; Nikolov, I.; Kruse, E.K.; Garnæs, J.; Madsen, C.B. Quantifying the Influence of Surface Texture and Shape on Structure from Motion 3D Reconstructions. *Sensors* **2023**, *23*. https://doi.org/10.3390/s23010178.
11. Ashida, K.; Santo, H.; Okura, F.; Matsushita, Y. Resolving Scale Ambiguity in Multi-view 3D Reconstruction Using Dual-Pixel Sensors. In Proceedings of the Computer Vision – ECCV 2024; Leonardis, A.; Ricci, E.; Roth, S.; Russakovsky, O.; Sattler, T.; Varol, G., Eds., Cham, 2025; pp. 162–178.
12. Zhang, Z. A flexible new technique for camera calibration. *IEEE Transactions on Pattern Analysis and Machine Intelligence* **2000**, *22*, 1330–1334. https://doi.org/10.1109/34.888718.
13. Hartley, R.I.; Sturm, P. Triangulation. *Computer Vision and Image Understanding* **1997**, *68*, 146–157. https://doi.org/https://doi.org/10.1006/cviu.1997.0547.
14. Marr, D.; Poggio, T.; Hildreth, E.C.; Grimson, W.E.L., A computational theory of human stereo vision. In *From the Retina to the Neocortex: Selected Papers of David Marr*; Birkhäuser Boston: Boston, MA, 1991; pp. 263–295. https://doi.org/10.1007/978-1-4684-6775-8_11.
15. Brown, D. Close-Range Camera Calibration. In Proceedings of the Proceedings of the Conference on Close-Range Camera Calibration, 1971.
16. Kannala, J.; Brandt, S. A generic camera calibration method for fish-eye lenses. In Proceedings of the Proceedings of the 17th International Conference on Pattern Recognition, 2004. ICPR 2004., 2004, Vol. 1, pp. 10–13 Vol.1. https://doi.org/10.1109/ICPR.2004.1333993.
17. Ma, L.; Chen, Y.; Moore, K.L. Rational Radial Distortion Models with Analytical Undistortion Formulae. *CoRR* **2003**, *cs.CV/0307047*.
18. Liao, K.; Nie, L.; Huang, S.; Lin, C.; Zhang, J.; Zhao, Y.; Gabbouj, M.; Tao, D. Deep learning for camera calibration and beyond: A survey. *arXiv preprint arXiv:2303.10559* **2023**.
19. Schops, T.; Larsson, V.; Pollefeys, M.; Sattler, T. Why Having 10,000 Parameters in Your Camera Model Is Better Than Twelve. In Proceedings of the Proceedings of the IEEE/CVF Conference on Computer Vision and Pattern Recognition (CVPR), June 2020.
20. Lopez, M.; Mari, R.; Gargallo, P.; Kuang, Y.; Gonzalez-Jimenez, J.; Haro, G. Deep single image camera calibration with radial distortion. In Proceedings of the Proceedings of the IEEE/CVF Conference on Computer Vision and Pattern Recognition, 2019, pp. 11817–11825.
21. Bogdan, O.; Eckstein, V.; Rameau, F.; Bazin, J.C. DeepCalib: a deep learning approach for automatic intrinsic calibration of wide field-of-view cameras. In Proceedings of the Proceedings of the 15th ACM SIGGRAPH European Conference on Visual Media Production, New York, NY, USA, 2018; CVMP '18. https://doi.org/10.1145/3278471.3278479.
22. Faig, W. CALIBRATION OF CLOSE-RANGE PHOTOGRAMMETRIC SYSTEMS: MATHEMATICAL FORMULATION. *Photogrammetric Engineering and Remote Sensing* **1975**, *41*.
23. Abdel-Aziz, Y.I.; Karara, H.M. Direct linear transformation from comparator coordinates into object space coordinates in close-range photogrammetry. In Proceedings of the Symposium on Close-Range Photogrammetry, Falls Church, VA, 1971.

24. Tsai, R. A versatile camera calibration technique for high-accuracy 3D machine vision metrology using off-the-shelf TV cameras and lenses. *IEEE Journal on Robotics and Automation* **1987**, *3*, 323–344. https: //doi.org/10.1109/JRA.1987.1087109.
25. Faugeras, O.D.; Luong, Q.T.; Maybank, S.J. Camera self-calibration: Theory and experiments. In Proceedings of the Computer Vision — ECCV'92; Sandini, G., Ed., Berlin, Heidelberg, 1992; pp. 321–334.
26. Maybank, S.J.; Faugeras, O.D. A theory of self-calibration of a moving camera. *International Journal of Computer Vision* **1992**, *8*, 123–151. https://doi.org/10.1007/BF00127171.
27. Li, X.; Wang, G.; Liu, J. Automatic camera calibration method based on dashed lines. In Proceedings of the Other Conferences, 2013.
28. Ma, S.D. A self-calibration technique for active vision systems. *IEEE Transactions on Robotics and Automation* **1996**, *12*, 114–120. https://doi.org/10.1109/70.481755.
29. Zhan, H. A Review on Some Active Vision Based Camera Calibration Techniques. *Chinese Journal of Computers* **2002**.
30. Kong, X.F.; Chen, Q.; Gu, G.H.; Qian, W.X.; Ren, K.; Wang, J.J. A GPS-based camera calibration method. *China Ordnance Society* **XXXX**. Publication details needed: year, volume, issue, pages.
31. Xiao, Z.; Jin, L.; Yu, D.; Tang, Z. A cross-target-based accurate calibration method of binocular stereo systems with large-scale field-of-view. *Measurement* **2010**, *43*, 747–754. https://doi.org/https://doi.org/10.1016/j. measurement.2010.01.017.
32. Shang, Y.; Sun, X.; Yang, X.; Wang, X.; Yu, Q. A camera calibration method for large field optical measurement. *Optik* **2013**, *124*, 6553–6558. https://doi.org/https://doi.org/10.1016/j.ijleo.2013.05.121.
33. Wang, X.; Gao, Y.; Wei, Z. Calibration method for large-field-of-view stereo vision system based on distancerelated distortion model. *Opt. Express* **2023**, *31*, 21816–21833. https://doi.org/10.1364/OE.492498.
34. Wang, Y.; Chen, X.; Tao, J.; Wang, K.; Ma, M. Accurate feature detection for out-of-focus camera calibration. *Appl. Opt.* **2016**, *55*, 7964–7971. https://doi.org/10.1364/AO.55.007964.
35. Wang, Y.; Liu, L.; Cai, B.; Wang, K.; Chen, X.; Wang, Y.; Tao, B. Stereo calibration with absolute phase target. *Opt. Express* **2019**, *27*, 22254–22267. https://doi.org/10.1364/OE.27.022254.
36. Najman, P.; Zemčík, P. Stereo camera pair calibration for traffic surveillance applications. *Optical Engineering* **2022**, *61*, 114103. https://doi.org/10.1117/1.OE.61.11.114103.
37. Wells, L.A.; Chung, W. Vision-Aided Localization and Mapping in Forested Environments Using Stereo Images. *Sensors* **2023**, *23*, 7043. https://doi.org/10.3390/S23167043.
38. Tang, Y.; Nie, Y.; Chen, Y.; Wang, Y. Real-time recognition and localization method for deep-sea underwater object. In Proceedings of the 2023 3rd International Conference on Neural Networks, Information and Communication Engineering (NNICE), 2023, pp. 276–281. https://doi.org/10.1109/NNICE58320.2023.1010 5685.
39. Zheng, S.; Liu, Y.; Weng, W.; Jia, X.; Yu, S.; Wu, Z. Tomato Recognition and Localization Method Based on Improved YOLOv5n-seg Model and Binocular Stereo Vision. *Agronomy* **2023**, *13*. https://doi.org/10.3390/ agronomy13092339.
40. Wang, Y.; Han, J.; Cao, C.; Li, F. Identification and Localization of Potato Bud Eye Based on Binocular Vision Technology. In Proceedings of the Frontier Computing; Hung, J.C.; Yen, N.Y.; Chang, J.W., Eds., Singapore, 2022; pp. 604–611.
41. Zheng, Y.; Liu, P.; Qian, L.; Qin, S.; Liu, X.; Ma, Y.; Cheng, G. Recognition and Depth Estimation of Ships Based on Binocular Stereo Vision. *Journal of Marine Science and Engineering* **2022**, *10*. https://doi.org/10.339 0/jmse10081153.
42. Liu, Y.; Bai, J.; Wang, G.; Wu, X.; Sun, F.; Guo, Z.; Geng, H. UAV Localization in Low-Altitude GNSSDenied Environments Based on POI and Store Signage Text Matching in UAV Images. *Drones* **2023**, *7*. https://doi.org/10.3390/drones7070451.
43. Zhang, Y.; Yang, J.; Li, G.; Zhao, T.; Song, X.; Zhang, S.; Li, A.; Bian, H.; Li, J.; Zhang, M. Camera Calibration for Long-Distance Photogrammetry Using Unmanned Aerial Vehicles. *Journal of Sensors* **2022**, *2022*, 8573315.
44. Zhou, Z.; Hu, Z.; Li, N.; Lai, G. Enhancing vehicle localization by matching HD map with road marking detection. *Proceedings of the Institution of Mechanical Engineers, Part D: Journal of Automobile Engineering* **2024**, *238*, 4129–4141, [https://doi.org/10.1177/09544070231191156]. https://doi.org/10.1177/09544070231191156.
45. Jin, Y.; Ren, Y.; Song, T.; Jiang, Z.; Song, G. On-board Monocular Vision-based Ground Target Recognition and Localization System for a Flapping-wing Aerial Vehicle. In Proceedings of the Proceedings of the 2023 4th International Conference on Computing, Networks and Internet of Things, New York, NY, USA, 2023; CNIOT '23, p. 278–285. https://doi.org/10.1145/3603781.3603829.
46. Han, S.; Do, M.D.; Kim, M.; Cho, S.; Choi, S.K.; Choi, H.J. A precise 3D scanning method using stereo vision with multipoint markers for rapid workpiece localization. *Journal of Mechanical Science and Technology* **2022**, *36*, 6307–6318. https://doi.org/10.1007/s12206-022-1142-2.
47. Punna, R.; Aravamuthan, G.; Kar, S. Recovering 3D from 2D Image Points (Calibrated Camera Approach for Surveillance). In Proceedings of the 2022 IEEE 7th International conference for Convergence in Technology (I2CT), 2022, pp. 1–7. https://doi.org/10.1109/I2CT54291.2022.9825444.

48. Pinggera, P.; Pfeiffer, D.; Franke, U.; Mester, R. Know Your Limits: Accuracy of Long Range Stereoscopic Object Measurements in Practice. In Proceedings of the Computer Vision – ECCV 2014; Fleet, D.; Pajdla, T.; Schiele, B.; Tuytelaars, T., Eds., Cham, 2014; pp. 96–111.
49. Reprojection error:, howpublished = https://support.pix4d.com/hc/en-us/articles/202559369#:~:text= Image%3A%20Pix4Dmatic_reprojection_error,.
50. Gavin, H.P. The Levenberg-Marquardt method for nonlinear least squares curve-fitting problems ©. 2013.
51. Opencv:Camera Calibration and 3D Reconstruction, howpublished = https://docs.opencv.org/4.x/d9/d0c/ group__calib3d.html,.
52. World Geodesic System:, howpublished = https://en.wikipedia.org/wiki/World_Geodetic_System,.
53. Paszke, A.; Gross, S.; Massa, F.; Lerer, A.; Bradbury, J.; Chanan, G.; Killeen, T.; Lin, Z.; Gimelshein, N.; Antiga, L.; et al. PyTorch: An Imperative Style, High-Performance Deep Learning Library, 2019, [arXiv:cs.LG/1912.01703].
54. Kingma, D.P.; Ba, J. Adam: A Method for Stochastic Optimization, 2017, [arXiv:cs.LG/1412.6980]. Taylor, J.; Wang, W.; Bala, B.; Bednarz, T. Optimizing the optimizer for data driven deep neural networks and physics informed neural networks, 2022, [arXiv:cs.LG/2205.07430]